\documentclass[10pt]{article}
\usepackage{hmi-preprint}
\usepackage{amsmath}
\usepackage{url}
\usepackage{booktabs}
\usepackage{nicefrac}
\usepackage{microtype}
\usepackage{marvosym}
\usepackage{xcolor}         
\definecolor{cvprblue}{rgb}{0.21,0.49,0.74}
\hypersetup{
    colorlinks=true, 
    linkcolor=cvprblue, 
    citecolor=cvprblue, 
    urlcolor=cvprblue,  
}

\usepackage{multirow}
\usepackage{graphicx}
\usepackage{caption}
\usepackage{placeins}

\newcommand{\method}{MemoryVLN}
\newcommand{\ind}{\mathbf{1}}
\newcommand{\appref}[1]{\hyperref[#1]{Appendix~\ref*{#1}}}

\preprintlogos{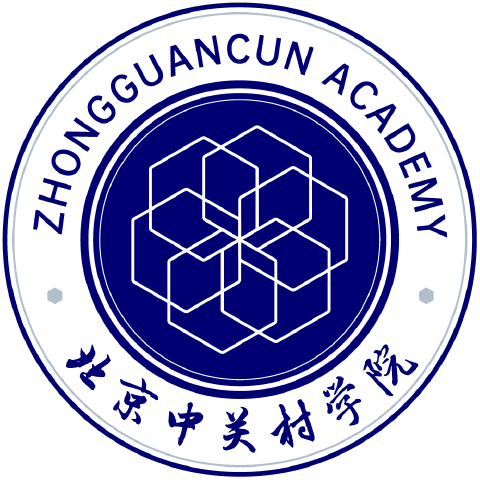}{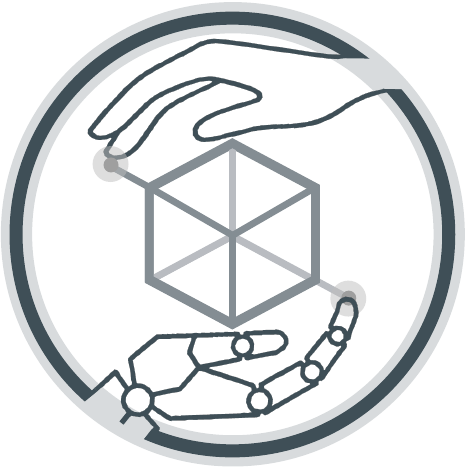}

\preprintdate{October 2026}

\runningtitle{MemoryVLN: From Ego-Aware Memory Representation to Cross-Context Memory Reuse in Vision-Language Navigation}

\preprintfooter{\textsuperscript{\Letter} Corresponding author.}

\preprintlinkset
  {https://zgca-hmi-lab.github.io/MemVLN}
  {https://github.com/ZGCA-HMI-Lab/MemVLN}{https://huggingface.co/xiaoxiaoxiaoyyy/MemVLN}

\preprintteaser[fig:overview]{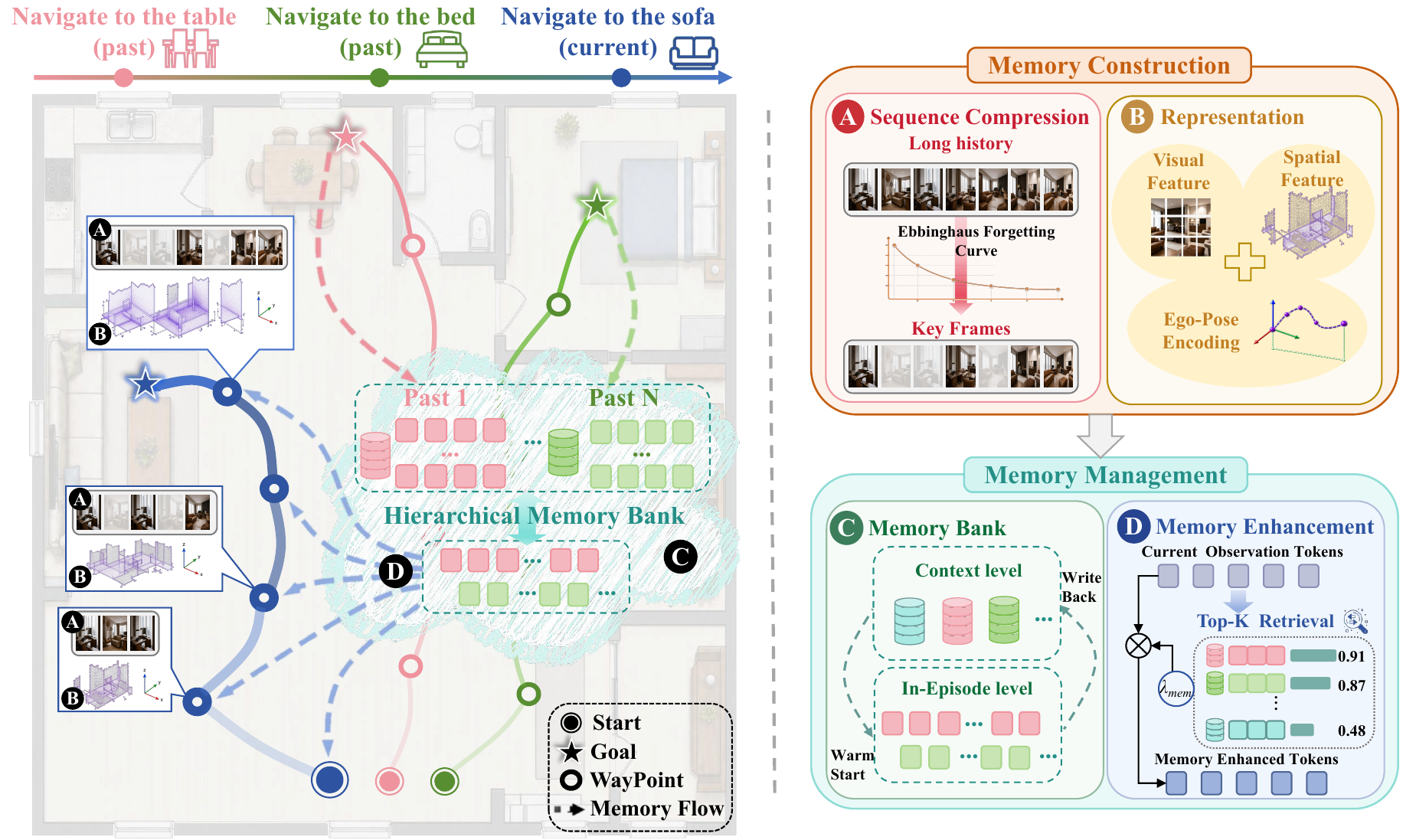}{Conceptual overview of MemoryVLN, comprising two stages:
    Memory Construction and Memory Management.
    Memory Construction transforms navigation histories into compact,
    spatially grounded memory tokens, while Memory Management maintains,
    retrieves, and reuses in-episode and cross-context experience.}

\title{MemoryVLN: From Ego-Aware Memory Representation to Cross-Context Memory Reuse in Vision-Language Navigation}
\author{
Xueyao Wei\textsuperscript{1,2} \hspace{0.15in}
Yifei Wang\textsuperscript{3} \hspace{0.15in}
Delun Li\textsuperscript{1,2} \hspace{0.15in}
Shaoan Wang\textsuperscript{3} \\[0.1in]
Junzhi Yu\textsuperscript{3} \hspace{0.15in}
Wenjie Song\textsuperscript{1} \hspace{0.15in}
Xingyu Chen\textsuperscript{2,\Letter}
}

\affil{
\textsuperscript{1}Beijing Institute of Technology \hspace{0.15in} 
\textsuperscript{2}Zhongguancun Academy
\hspace{0.15in} 
\textsuperscript{3}Peking University
}

\begin{document}

\maketitle

\begin{abstract}
Large vision-language models (LVLMs) have shown strong potential for embodied vision-language navigation (VLN), yet their performance degrades on long-horizon trajectories due to context-length limits, redundant historical observations, and unstable spatial alignment across time. These issues make it difficult to both compress long navigation histories and reliably reuse past spatial experience. To address these limitations, we propose \textbf{MemoryVLN}, a unified memory-driven navigation system built around memory construction and management. During memory construction, \emph{Long Sequence Compression} employs forgetting-curve-inspired temporal sampling to preserve dense recent evidence and sparse yet representative distant observations under a limited token budget. \emph{Ego-Aware Memory Representation} further integrates 2D semantic tokens, 3D geometric tokens, and 6D agent poses, allowing each memory to encode both what was observed and where it was observed. During memory management, \emph{Hierarchical Memory Bank} maintains in-episode and context-level memories through cross-context warm-start and confidence-controlled write-back. \emph{Memory Enhancement} then retrieves the Top-$K$ relevant memories according to the current navigation state and fuses them with the current visual tokens. Together, these mechanisms transform raw navigation history into compact, spatially grounded, selectively retrievable, and reusable navigation experience. Extensive experiments on R2R-CE and RxR-CE demonstrate consistent improvements over the state-of-the-art baselines by a substantial margin (i.e., 2\% and 6.5\% in successful rate). 
\end{abstract}

\section{Introduction}

Traditional VLN and VLN-CE formulations usually treat navigation as a sequential decision-making problem within a single instruction episode~\citep{anderson2018r2r,ku2020rxr,krantz2020vlnce,jain2019r4r,qi2020reverie,thomason2020cvdn}. The agent predicts actions from language instructions and current visual observations, and most existing methods focus on vision-language alignment, action prediction, and local history modeling. This formulation is effective for short contexts, but it does not fully capture the difficulty of real long-horizon navigation.


Real long-horizon navigation is not merely a scaled-up short-context problem. The agent should remember landmarks many steps earlier, previously visited paths, incorrect branches, and the surrounding spatial layout. As the trajectory becomes longer, observations accumulate rapidly, while critical evidence may appear far before the final decision. The core challenge is therefore not to see more frames, but to preserve the most useful historical experience within a limited context budget, echoing the distinction between short-term processing and persistent memory in cognitive models~\citep{atkinson1968humanmemory,baddeley1974workingmemory}.

Although recent LVLM-based VLN methods significantly improve instruction understanding and multimodal reasoning~\citep{zhou2023navgpt}, many of them encode accumulated observations as video or context sequences~\citep{navid2024,uninavid2024}, retain slow-fast contexts~\citep{streamvln2025}, summarize history through recurrent or latent states~\citep{hong2021vlnbert,chen2021hamt}, or adopt task-specific history aggregation and reuse mechanisms~\citep{reborn2023,janusvln2026}. They rarely address four coupled questions in a unified way: how to compress long histories, how to construct ego-aware representations of past observations, how to organize and update memories across contexts, and how to retrieve relevant experience for current navigation decisions. As a result, they remain vulnerable to redundant historical frames, context forgetting, spatial misalignment, and the inability to transfer useful experience across contexts.

In this paper, we propose \textbf{MemoryVLN}, a unified memory-driven navigation system organized around memory construction and memory management, as illustrated in \autoref{fig:overview}. During memory construction, \emph{Long Sequence Compression} combines inter-chunk budget allocation with intra-chunk key-frame selection to preserve dense recent evidence and sparse yet representative distant observations under a limited token budget. \emph{Ego-Aware Visuospatial Representation} then integrates ViT-based 2D visual tokens, VGGT-based 3D geometric tokens, and 6D agent poses to construct spatially grounded memory tokens. During memory management, \emph{Memory Bank} maintains in-episode experience through a FIFO short-term pool and a cosine-weighted long-term pool, while a context-level bank supports cross-context memory warm start and confidence-gated write-back. Finally, \emph{Memory Enhancement} retrieves the Top-$K$ memories most relevant to the current state and residually fuses them with the current-frame tokens for action prediction. Together, these modules transform accumulated observations into compact, spatially grounded, selectively retrievable, and reusable navigation experience.

Our contributions are summarized as follows:

\begin{itemize}
    \item We propose \textbf{MemoryVLN}, a unified memory-driven navigation system organized around memory construction and memory management, covering the complete memory process of compression, representation, storage, update, retrieval, and reuse.

    \item We design four modules for the whole memory lifecycle: \emph{Long Sequence Compression} for reducing long-history redundancy under a fixed token budget; \emph{Ego-Aware Memory Representation} for constructing spatially grounded memories from 2D semantics, 3D geometry, and 6D agent poses; \emph{Hierarchical Memory Bank} for maintaining in-episode and context-level experience; and \emph{Memory Enhancement} for retrieving and fusing relevant memories.

    \item Experiments on R2R-CE and RxR-CE demonstrate that MemoryVLN improves the success rate over state-of-the-art baselines by approximately 2\% and 6.5\%, respectively, validating the effectiveness of our memory designs for long-horizon navigation.
\end{itemize}

\FloatBarrier
\section{Related Work}

\paragraph{Embodied Navigation.}
VLN has expanded from R2R to long-path, multilingual, dialog-based, and object-oriented tasks~\citep{anderson2018r2r,jain2019r4r,ku2020rxr,thomason2020cvdn,qi2020reverie}, primarily supported by Matterport3D and Habitat~\citep{chang2017matterport3d,savva2019habitat}. Most benchmarks evaluate episodes independently, whereas MemoryVLN retains long-horizon history and reuses experience across episodes.

Classical VLN improves navigation through data augmentation~\citep{fried2018speakerfollower,tan2019envdrop} and history or map encoding~\citep{hong2021vlnbert,chen2021hamt,pashevich2021episodic,chen2022duet,an2023bevbert}. Continuous methods improve execution through waypoint prediction, cross-environment transfer, and history reuse~\citep{krantz2020vlnce,hpn_dn2021,raychaudhuri2021law,sim2sim2022,cma2022,reborn2023}. However, they typically rely on bounded histories, explicit maps, or episode-specific planning rather than a unified memory that can be compressed, spatially grounded, and reused.

LLM- and LVLM-based agents strengthen reasoning and action generation~\citep{zhou2023navgpt,navid2024,uninavid2024,navila2024}, while navigation foundation models improve context modeling~\citep{streamvln2025,internvla_n1_2025,janusvln2026}. World models support planning through future prediction~\citep{koh2021pathdreamer,dreamwalker2023,yao2025navmorph,futurenav2026}, whereas MemoryVLN covers memory compression, representation, storage, retrieval, and update without additional navigation data.

\paragraph{Memory for Long-Horizon and Persistent Navigation.}
Cognitive models characterize short-term, working, and long-term storage and temporal decay~\citep{ebbinghaus1885memory,atkinson1968humanmemory,baddeley1974workingmemory}, while Memory Networks provide addressable external memory~\citep{weston2015memorynetworks}. This hierarchy suggests that a navigation agent should preserve detailed recent evidence while consolidating stable information into longer-term experience. However, general memory mechanisms do not directly address visual redundancy, pose misalignment, or unreliable write-back in navigation. Navigation memory must additionally decide what to retain, compress, retrieve, or transfer as observations accumulate online.

Navigation memory has used history-aware and episodic encoders, trajectory reuse, spatial maps, selective storage, and continual retrieval~\citep{chen2021hamt,pashevich2021episodic,reborn2023,gridmm2023,mapnav2025,cosmo2025,janusvln2026,li2026cmmrvln}; Iterative VLN additionally retains experience across ordered episodes~\citep{krantz2023ivln}. Most methods still center on single episodes, fixed implicit states, or static maps. MemoryVLN instead combines short- and long-term pools with a persistent Context-Level Bank, thresholded retrieval, warm start, confidence gating, and LRU replacement.

\paragraph{Spatial and Geometric Representations for Navigation.}
Semantic maps and BEV or grid representations explicitly organize landmarks and spatial relations to improve navigation reasoning~\citep{ego2map2023,an2023bevbert,gridmm2023}. 3D feature fields, RGB-D video models, and dual implicit memories further capture cross-view geometry~\citep{g3dlf2025,sim2real2024,navid4d2025,janusvln2026}. However, these methods often rely on depth or preconstructed maps and mainly support episode-specific localization or planning, rather than maintaining geometric tokens that can be compressed, retrieved, and updated across contexts. MemoryVLN instead binds visual appearance, geometry, and pose within retrievable memory tokens, enabling historical observations to support the current decision.

DUSt3R and VGGT provide strong geometric priors through cross-view matching and reconstruction~\citep{wang2024dust3r,wang2025vggt}. Yet they are perception backbones and do not determine which observations should survive long trajectories or how experience should be reused across contexts. MemoryVLN fuses frozen VGGT geometry with 2D visual tokens and 6D poses, then stores the constructed tokens in an updatable hierarchical memory bank.

\FloatBarrier
\section{Method}

\subsection{Problem Formulation}

In VLN-CE, an agent receives an instruction \(I\) and a stream of egocentric RGB observations \(\{o_t\}_{t=1}^{T}\). At each step \(t\), it selects an action from a discrete set containing \textsc{MoveForward}, \textsc{TurnLeft}, \textsc{TurnRight}, and \textsc{Stop}. Long-horizon decisions require not only the current observation but also evidence accumulated within the trajectory and experience retained from previous episodes in the same environment. Instead of repeatedly encoding the complete history, MemoryVLN organizes it into an In-Episode Bank \(\mathcal{B}^{\mathrm{epi}}_t\) and a persistent Context-Level Bank \(\mathcal{B}^{\mathrm{ctx}}\). The navigation policy is therefore defined as
\begin{equation}
    a_t \sim p_{\theta}\!\left(
        a_t \mid I,o_{\leq t},
        \mathcal{B}^{\mathrm{epi}}_t,
        \mathcal{B}^{\mathrm{ctx}}
    \right).
\end{equation}
Here, \(\mathcal{B}^{\mathrm{epi}}_t\) contains the short- and long-term pools available at step \(t\), while \(\mathcal{B}^{\mathrm{ctx}}\) provides persistent cross-episode experience. This decomposition separates transient task evidence from reusable environmental priors.

\subsection{Overview}

\begin{figure*}[!t]
    \centering
    \includegraphics[width=0.85\textwidth]{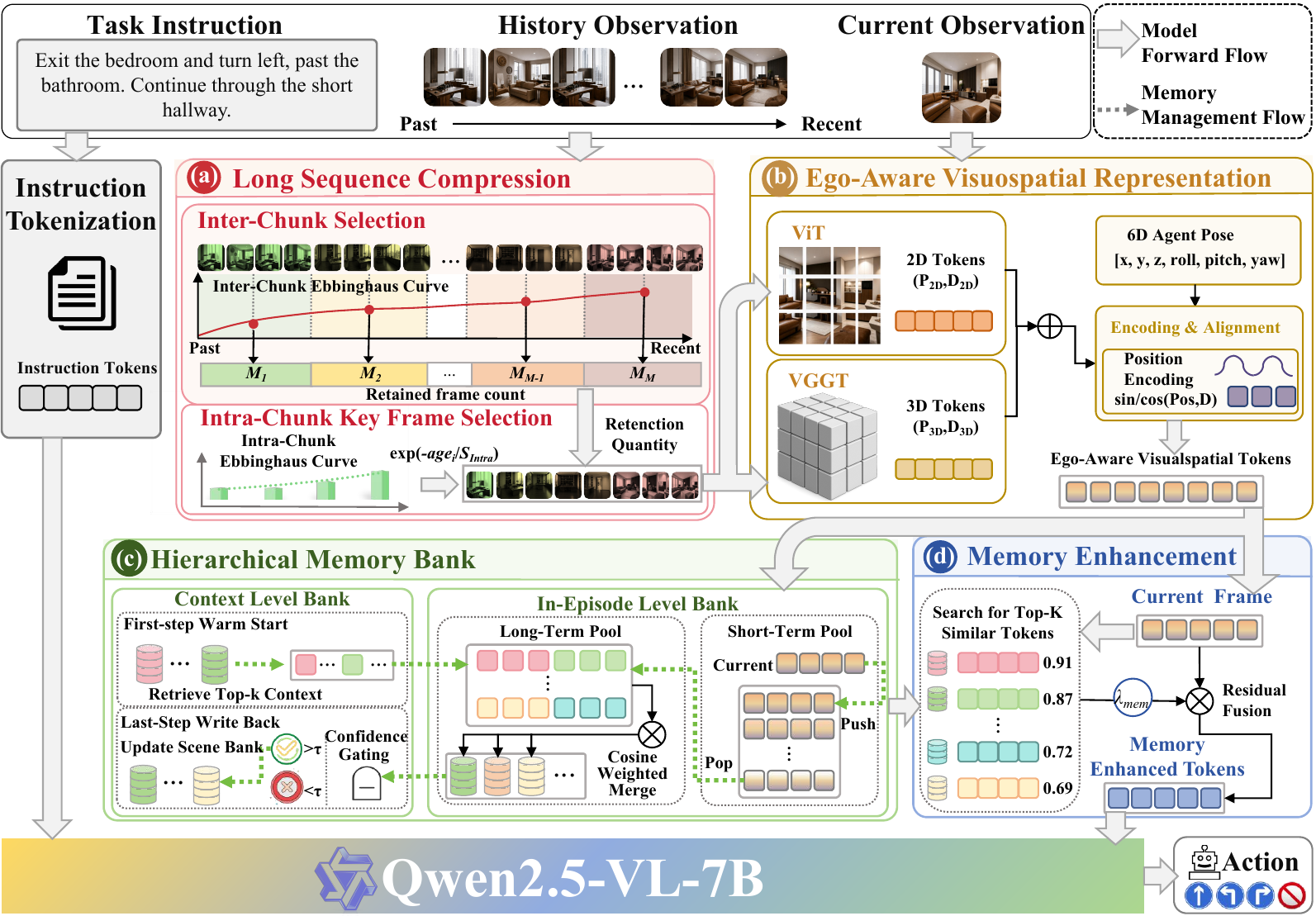}
    \caption{Detailed pipeline of MemoryVLN. (a) \emph{Long Sequence Compression} performs inter-chunk budget allocation and intra-chunk key-frame selection. (b) \emph{Ego-Aware Visuospatial Representation} integrates ViT-based 2D tokens, VGGT-based 3D tokens, and 6D agent poses. (c) \emph{Memory Bank} maintains context-level and in-episode memories through warm start, write-back, and short-/long-term memory updates. (d) \emph{Memory Enhancement} retrieves Top-$K$ relevant memories and residually fuse them with the current-frame tokens for action prediction.}
    \label{fig:method}
\end{figure*}
\autoref{fig:overview} and~\autoref{fig:method} show the conceptual pipeline and implementation details of MemoryVLN. The framework contains two stages, \emph{Memory Construction} and \emph{Memory Management}, implemented by four modules corresponding to panels (a)-(d). During construction, Long Sequence Compression bounds the visual-token budget, and Ego-Aware Visuospatial Representation attaches appearance, geometry, and pose information to the retained observations. During management, Hierarchical Memory Bank organizes in-episode and cross-episode experience, while Memory Enhancement retrieves history relevant to the current state. Instruction tokens remain intact throughout the trajectory; only the growing visual history is selectively compressed. The resulting memory-enhanced tokens are passed with the instruction to the Transformer-based Qwen2.5-VL backbone for action prediction~\citep{vaswani2017attention,bai2025qwen25vl}. In this way, Memory Construction converts raw observations into structured memory, while Memory Management determines when the memory is stored, retrieved, and updated.

\subsection{Long Sequence Compression}

\paragraph{Adaptive Retention Budget.}
Retaining every historical frame makes the number of visual tokens grow linearly with trajectory length. Long Sequence Compression instead uses an Ebbinghaus-inspired retention prior~\citep{ebbinghaus1885memory} to preserve dense recent evidence and sparse representative distant observations under a bounded budget. For a history of \(N\) frames, the target retention budget is
\begin{equation}
    B(N)=\min\!\left\{
        N,\max\!\left[B_{\min},
        \left\lfloor a+b\sqrt{N}\right\rfloor\right]
    \right\}.
\end{equation}
Here, \(B_{\min}\) is the minimum budget and \(a,b\) control its sublinear growth; an optional frame cap further limits evaluation-time memory. The square-root schedule avoids abruptly discarding early landmarks while preventing the visual context from growing without bound.

\paragraph{Inter-Chunk Selection.}
The history is first divided into \(C\) consecutive chunks \(\{\mathcal{C}_i\}_{i=1}^{C}\) so that the budget can be distributed across temporal ranges. For chunk \(i\), its age is \(\mathrm{age}_i=C-i\), and its retention weight is \(w_i=\exp(-\mathrm{age}_i/S_{\mathrm{inter}})\). After normalization by \(\hat{w}_i=w_i/\sum_{j=1}^{C}w_j\), the initial allocation is
\begin{equation}
    M_i^{(0)}=\min\!\left\{
        |\mathcal{C}_i|,
        \left\lfloor B(N)\hat{w}_i\right\rfloor
    \right\}.
\end{equation}
More recent chunks therefore receive larger weights and denser coverage. Unused slots are assigned by the largest fractional remainders among unsaturated chunks, producing dense recent coverage while retaining selected long-range evidence.

\paragraph{Intra-Chunk Key-Frame Selection.}
Within chunk \(\mathcal{C}_i\), frame \(f_{i,k}\) is sampled without replacement according to
\begin{equation}
    p_{i,k}=
    \frac{\exp(-\mathrm{age}_{i,k}/S_{\mathrm{intra}})}
         {\sum_{\ell}\exp(-\mathrm{age}_{i,\ell}/S_{\mathrm{intra}})}.
\end{equation}
Unlike deterministic top-\(M_i\) selection, probabilistic sampling reduces clusters of near-duplicate frames and improves temporal diversity. The most recent frame of every sampled chunk is reserved, and the first and last frames of the complete trajectory are explicitly enforced through \(\mathcal{H}^{\mathrm{cmp}}=\mathrm{EnsureFirstLast}(\bigcup_i\mathrm{Sample}(\mathcal{C}_i,M_i))\). This two-level selection retains local continuity and distant landmarks without linearly increasing visual-token cost.

\subsection{Ego-Aware Visuospatial Representation}

\paragraph{2D and 3D Token Extraction.}
This module extracts complementary appearance and geometry from the retained observations. The Qwen2.5-VL ViT produces frame-wise 2D appearance tokens~\citep{bai2025qwen25vl}, while the frozen VGGT sequentially aggregates retained frames through cached historical states and outputs geometry-oriented tokens from its final state~\citep{wang2025vggt}. Separate patch mergers project both streams to the Qwen2.5-VL hidden dimension, after which the VGGT tokens are repeated and truncated to match the 2D-token count. The 2D stream preserves local semantic detail, whereas the 3D stream supplies structural constraints across viewpoints.

\paragraph{6D Ego-Pose Encoding and Alignment.}
Visual appearance alone can confuse repeated landmarks or similar rooms observed from different locations. Each retained frame is therefore associated with the relative 6D pose \(\mathbf{p}_t=[x_t,y_t,z_t,\mathrm{roll}_t,\mathrm{pitch}_t,\mathrm{yaw}_t]\). Positions are normalized by a spatial scale, while Euler angles are wrapped to \([-\pi,\pi]\), normalized separately, and mapped by a fixed sinusoidal projector \(\phi(\cdot)\). Before patch merging, the ViT branch injects the pose of each retained frame as \(\mathbf{Z}^{2D}_t=\mathrm{Merge}_{2D}(\mathbf{T}^{2D}_t+\mathcal{B}(\phi(\mathbf{p}_t)))\). The final VGGT state instead uses the latest retained pose, giving \(\mathbf{Z}^{3D}=\mathrm{Merge}_{3D}(\mathbf{T}^{3D}_{1:F}+\mathcal{B}(\phi(\mathbf{p}_F)))\). Here, \(\mathcal{B}(\cdot)\) broadcasts the pose embedding to all relevant tokens.

\paragraph{Token Alignment and Visuospatial Fusion.}
The frame-wise tokens are concatenated as \(\mathbf{Z}^{2D}=\mathrm{Concat}_{t=1}^{F}\mathbf{Z}^{2D}_t\), and the final VGGT tokens are repeated and truncated by \(\mathrm{Align}(\cdot)\). The aligned streams are fused through a geometry residual:
\begin{equation}
    \mathbf{V}=\mathbf{Z}^{2D}
    +\lambda_{\mathrm{geo}}\,
      \mathrm{Align}(\mathbf{Z}^{3D}),
\end{equation}
where \(\lambda_{\mathrm{geo}}\) controls the contribution of the geometric branch. This formulation keeps the pretrained ViT representation as the primary stream instead of replacing it with geometry. The resulting tokens jointly encode visual content, observation pose, and structural relations across historical viewpoints.

\subsection{Hierarchical Memory Bank}

\paragraph{Context-Level Bank and Warm Start.}
The Context-Level Bank in \autoref{fig:method}(c) persistently stores reliable experience across episodes in the same environment. A context key combines mean-pooled visual evidence \(\mathbf{k}^{v}\) and an optional instruction representation \(\mathbf{k}^{I}\):
\begin{equation}
    \mathbf{k}=\mathrm{Norm}\!\left(
        (1-\beta)\mathrm{Norm}(\mathbf{k}^{v})
        +\beta\mathrm{Norm}(\mathbf{k}^{I})
    \right).
\end{equation}
At the beginning of a new episode, the top-ranked entry is retrieved by cosine similarity and accepted only when its score exceeds a context threshold, preventing unrelated priors from entering the current state. For an accepted prior \(\mathbf{c}^{\star}\), warm start is performed as
\begin{equation}
    \mathbf{h}_0=(1-\tau_w)\mathbf{x}_0
    +\tau_w\mathbf{c}^{\star},
\end{equation}
and the resulting offset is broadcast to the first-frame tokens. This initialization allows the agent to reuse previously observed structure instead of rebuilding its memory.

\paragraph{In-Episode Level Bank.}
The In-Episode Bank contains a FIFO short-term pool \(\mathcal{M}_s\) and a capacity-limited long-term pool \(\mathcal{M}_l\). Recent frame patches remain intact in \(\mathcal{M}_s\), supporting local action continuity and short-range correction. When the short-term pool is full, its oldest entry is promoted to \(\mathcal{M}_l\), extending the lifetime of evidence that may become useful later. Once the long-term pool reaches capacity, an incoming entry is merged into its top-\(k\) most similar slots using coefficients adapted to cosine similarity and slot age, corresponding to the \emph{Cosine Weighted Merge} in \autoref{fig:method}. This dynamic transfer preserves recent detail while compressing distant experience.

\paragraph{Confidence-Gated Context Write-Back.}
At the final step, the system constructs a candidate context key and value from the aggregated in-episode memory. Its reliability is summarized by \(\gamma=\frac{1}{2}[\mathrm{Mean}(\mathbf{c}_s)+\mathrm{Mean}(\mathbf{c}_l)]\), where \(\mathbf{c}_s\) and \(\mathbf{c}_l\) are the confidence values of the short- and long-term pools. The candidate is written back only when \(\gamma\geq\delta\); this gate reduces persistent contamination from uncertain trajectories, while disabling confidence gating directly writes each valid pair. If the Context-Level Bank is full, the least-recently-used entry is replaced.

\subsection{Memory Enhancement}

\paragraph{Top-\ensuremath{k} Retrieval and Residual Fusion.}
Average pooling first converts the current-frame tokens into a query \(\mathbf{q}_t\), which retrieves the top-\(k\) entries from \(\mathcal{M}_s\cup\mathcal{M}_l\). Their weights are computed as \(\alpha_{t,j}=\mathrm{softmax}_{j}(\mathrm{cos}(\mathbf{q}_t,\mathbf{m}_j)/\tau_m)\), yielding the summary \(\bar{\mathbf{m}}_t=\sum_{j\in\mathcal{N}_k(t)}\alpha_{t,j}\mathbf{m}_j\). We then interpolate at the frame level using \(\mathbf{v}^{\mathrm{mem}}_t=(1-\rho)\mathbf{q}_t+\rho\bar{\mathbf{m}}_t\), and broadcast its offset to all patch tokens of frame \(t\). The final memory-enhanced representation is
\begin{equation}
    \widetilde{\mathbf{V}}
    =\mathbf{V}+\lambda_{\mathrm{mem}}
      (\mathbf{V}^{\mathrm{mem}}-\mathbf{V}).
\end{equation}
This two-stage fusion keeps the current observation as the primary signal while injecting distant evidence when it is relevant.

\subsection{Training and Inference}

MemoryVLN is trained for language-conditioned action prediction using standard VLN trajectories, without DAgger rollouts or additional navigation data. Qwen2.5-VL serves as the action-prediction backbone, while the VGGT geometry branch remains frozen. Training optimizes the task-specific visual projections, memory fusion, and action-prediction components jointly with the navigation objective. During closed-loop inference, compression, pose injection, hierarchical memory maintenance, top-\(k\) enhancement, and confidence-gated write-back are executed online. The In-Episode Bank is reset for each episode, whereas the Context-Level Bank persists within the corresponding environment scope. This asymmetric lifecycle prevents transient action noise from propagating across episodes while allowing reliable environmental experience to accumulate over time.
\section{Experiments}

\subsection{Experimental Setup}

\paragraph{Datasets and Training.}
We evaluate MemoryVLN on the R2R-CE and RxR-CE continuous-navigation benchmarks~\citep{anderson2018r2r,ku2020rxr,krantz2020vlnce}, built on Matterport3D environments and the Habitat simulator~\citep{chang2017matterport3d,savva2019habitat}. The model uses only standard VLN training trajectories, without DAgger rollouts~\citep{ross2011dagger} or additional navigation data. Qwen2.5-VL serves as the action-prediction backbone~\citep{bai2025qwen25vl}, while the VGGT geometry branch remains frozen during training~\citep{wang2025vggt}. Detailed training configurations, hyperparameters, and implementation settings are provided in the supplementary material.

\paragraph{Metrics.}
We report five standard VLN metrics: Navigation Error (NE), Oracle Success Rate (OS), Success Rate (SR), Success weighted by Path Length (SPL), and normalized Dynamic Time Warping (nDTW)~\citep{anderson2018evaluation,ku2020rxr}. NE measures the final distance to the goal, while OS and SR measure goal visitation and final-task success, respectively. SPL evaluates success together with path efficiency, and nDTW measures the consistency between predicted and reference trajectories.

\paragraph{Implementation Details.}
Unless otherwise specified, all four core modules are enabled: Long Sequence Compression, Ego-Aware Visuospatial Representation, Memory Bank, and Memory Enhancement. The Memory Bank contains in-episode short- and long-term pools together with a persistent Context-Level Bank. Frame selection, pose injection, memory retrieval, updates, context warm start, and confidence-gated write-back are performed online during navigation.

\subsection{Main Results}

\autoref{tab:main_comparison} compares MemoryVLN with representative VLN methods on the R2R-CE and RxR-CE val-unseen splits. MemoryVLN does not use DAgger rollouts or additional navigation data, whereas the baseline results follow their respective published training and evaluation settings. 
In particular, methods trained with large-scale cross-task or additional navigation data are not fully data-matched to MemoryVLN.

MemoryVLN achieves consistently strong performance on both benchmarks. On R2R-CE, it substantially reduces navigation error while achieving strong SR and SPL. This result demonstrates improvements in both navigation success and path efficiency. On the longer and more linguistically complex RxR-CE benchmark, MemoryVLN maintains a larger advantage in success rate. Compared with the strongest data-comparable baselines, MemoryVLN improves SR by approximately 2\% on R2R-CE and 6.5\% on RxR-CE.

These results indicate that the improvements do not rely on additional navigation data. Instead, they primarily arise from compressing redundant history, constructing ego-aware spatial memories, maintaining multi-timescale experience, and retrieving relevant historical evidence for current decisions.

\subsection{Ablation Study}

\autoref{tab:memory_ablation} reports ablations on the R2R-CE and RxR-CE val-unseen splits under identical settings. We remove Long Sequence Compression, pose encoding, the In-Episode Bank, and the Context-Level Bank individually. The full model performs best on all R2R-CE metrics and on SR, SPL, and nDTW for RxR-CE. Overall, the four modules provide complementary benefits in scalability, spatial alignment, in-episode history preservation, and cross-episode reuse.
\begin{table*}[!t]
\centering
\small
\setlength{\tabcolsep}{1.5pt}
\caption{Comparison with SOTA methods on R2R-CE and RxR-CE val-unseen split.  $^*$ denotes methods using a waypoint predictor, while $\dagger$ denotes models trained only with VLN-CE expert trajectories without additional navigation data. Pano, Odo, Depth, and S.RGB respectively represent panoramic view, odometry, depth, and single RGB.}
\label{tab:main_comparison}
\begin{tabular}{lcccc|cccc|cccc}
\toprule
\multirow{2}{*}{\textbf{Method}} & \multicolumn{4}{c|}{\textbf{Observation}} & \multicolumn{4}{c|}{\textbf{R2R Val-Unseen}} & \multicolumn{4}{c}{\textbf{RxR Val-Unseen}} \\
 & \textbf{Pano.} & \textbf{Odo.} & \textbf{Depth} & \textbf{S.RGB} & \textbf{NE$\downarrow$} & \textbf{OS$\uparrow$} & \textbf{SR$\uparrow$} & \textbf{SPL$\uparrow$} & \textbf{NE$\downarrow$} & \textbf{SR$\uparrow$} & \textbf{SPL$\uparrow$} & \textbf{nDTW$\uparrow$} \\
\midrule
HPN+DN$^{*}$~\citep{hpn_dn2021} & $\checkmark$ & $\checkmark$ & $\checkmark$ & -- & 6.31 & 40.0 & 36.0 & 34.0 & -- & -- & -- & -- \\
CMA$^{*}$~\citep{cma2022} & $\checkmark$ & $\checkmark$ & $\checkmark$ & -- & 6.20 & 52.0 & 41.0 & 36.0 & 8.76 & 26.5 & 22.1 & 47.0 \\
Sim2Sim$^{*}$~\citep{sim2sim2022} & $\checkmark$ & $\checkmark$ & $\checkmark$ & -- & 6.07 & 52.0 & 43.0 & 36.0 & -- & -- & -- & -- \\
VLN-BERT$^{*}$~\citep{hong2021vlnbert} & $\checkmark$ & $\checkmark$ & $\checkmark$ & -- & 5.74 & 53.0 & 44.0 & 39.0 & 8.98 & 27.0 & 22.6 & 46.7 \\
Ego$^2$-Map$^{*}$~\citep{ego2map2023} & $\checkmark$ & $\checkmark$ & $\checkmark$ & -- & 5.54 & 56.0 & 47.0 & 41.0 & -- & -- & -- & -- \\
DreamWalker$^{*}$~\citep{dreamwalker2023} & $\checkmark$ & $\checkmark$ & $\checkmark$ & -- & 5.53 & 59.0 & 49.0 & 44.0 & -- & -- & -- & -- \\
GridMM$^{*}$~\citep{gridmm2023} & $\checkmark$ & $\checkmark$ & $\checkmark$ & -- & 5.11 & 61.0 & 49.0 & 41.0 & -- & -- & -- & -- \\
Reborn$^{*}$~\citep{reborn2023} & $\checkmark$ & $\checkmark$ & $\checkmark$ & -- & 5.40 & 57.0 & 50.0 & 46.0 & 5.98 & 48.6 & 42.0 & 63.3 \\
InstructNav~\citep{instructnav2024} & $\checkmark$ & -- & -- & -- & 6.89 & -- & 31.0 & 24.0 & -- & -- & -- & -- \\
\midrule
COSMO~\citep{cosmo2025} & $\checkmark$ & -- & -- & -- & -- & 56.0 & 47.0 & 40.0 & -- & -- & -- & -- \\
AO-Planner~\citep{chen2025aoplanner} & $\checkmark$ & -- & -- & $\checkmark$ & 5.55 & 59.0 & 47.0 & 33.0 & 7.06 & 43.3 & 30.5 & 50.1 \\
LAW~\citep{raychaudhuri2021law} & -- & $\checkmark$ & $\checkmark$ & $\checkmark$ & 6.83 & 44.0 & 35.0 & 31.0 & 10.90 & 8.0 & 8.0 & 38.0 \\
MapNav~\citep{mapnav2025} & -- & $\checkmark$ & $\checkmark$ & $\checkmark$ & 4.93 & 53.0 & 39.7 & 37.2 & -- & -- & -- & -- \\
g3D-LF~\citep{g3dlf2025} & -- & $\checkmark$ & $\checkmark$ & $\checkmark$ & 5.70 & 59.5 & 47.2 & 34.6 & -- & -- & -- & -- \\
Seq2Seq~\citep{krantz2020vlnce} & -- & -- & $\checkmark$ & $\checkmark$ & 7.77 & 37.0 & 25.0 & 22.0 & 12.10 & 13.9 & 11.9 & 30.8 \\
NaVid-4D~\citep{navid4d2025} & -- & -- & $\checkmark$ & $\checkmark$ & 5.99 & 55.7 & 43.8 & 37.1 & -- & -- & -- & -- \\
NavMorph~\citep{yao2025navmorph} & -- & -- & $\checkmark$ & $\checkmark$ & 5.75 & 56.9 & 47.9 & 33.2 & 8.85 & 30.8 & 22.8 & 44.2 \\
\midrule
NaVid~\citep{navid2024} & -- & -- & -- & $\checkmark$ & 5.72 & 49.2 & 41.9 & 36.5 & 5.72 & 45.7 & 38.2 & -- \\
Sim2Real~\citep{sim2real2024} & -- & -- & -- & $\checkmark$ & 5.95 & 55.8 & 44.9 & 30.4 & 8.79 & 36.7 & 25.5 & 18.1 \\
Uni-NaVid~\citep{uninavid2024} & -- & -- & -- & $\checkmark$ & 5.58 & 53.3 & 47.0 & 42.7 & 6.24 & 48.7 & 40.9 & -- \\
StreamVLN-RGB-only~\citep{streamvln2025} & -- & -- & -- & $\checkmark$ & 5.10 & \textbf{64.0} & \textbf{55.7} & 50.9 & 6.16 & 51.8 & 45.0 & 62.1 \\
NaVILA~\citep{navila2024} & -- & -- & -- & $\checkmark$ & 5.22 & 62.5 & 54.0 & 49.0 & 6.77 & 49.3 & 44.0 & 58.8 \\
InternVLA-N1~\citep{internvla_n1_2025} & -- & -- & -- & $\checkmark$ & 4.89 & 60.6 & 55.4 & \textbf{52.1} & 6.41 & 49.5 & 41.8 & 62.6 \\
JanusVLN$^{\dagger}$~\citep{janusvln2026} & -- & -- & -- & $\checkmark$ & 5.17 & 58.0 & 52.8 & 49.2 & 6.46 & 51.4 & 44.3 & 59.1 \\
\midrule
\textbf{MemoryVLN (Ours)} & -- & -- & -- & $\checkmark$ & \textbf{3.52} & 59.98 & 54.76 & 50.26 & \textbf{3.33} & \textbf{57.84} & \textbf{50.42} & \textbf{63.52} \\
\bottomrule
\end{tabular}
\end{table*}
\FloatBarrier
\begin{table*}[!t]
    \centering
    \small
    \setlength{\tabcolsep}{4.5pt}
    \renewcommand{\arraystretch}{1.08}
    \caption{Component ablations on the R2R-CE and RxR-CE val-unseen splits.}
    \label{tab:memory_ablation}
    \begin{tabular}{@{}l ccc ccc@{}}
        \toprule
        & \multicolumn{3}{c}{\textbf{R2R-CE}}
        & \multicolumn{3}{c}{\textbf{RxR-CE}} \\
        \cmidrule(lr){2-4} \cmidrule(lr){5-7}
        \textbf{Variant}
        & \textbf{NE}$\downarrow$ & \textbf{SR}$\uparrow$ & \textbf{SPL}$\uparrow$
        & \textbf{NE}$\downarrow$ & \textbf{SR}$\uparrow$ & \textbf{SPL}$\uparrow$ \\
        \midrule
        w/o Long-Seq. Comp.
        & \multicolumn{3}{c}{\textit{OOM}}
        & \multicolumn{3}{c}{\textit{OOM}} \\
        w/o Pos-Encoding
        & 3.66 & 52.75 & 49.25
        & 3.35 & 57.24 & 48.74 \\
        w/o In-Episode
        & 3.57 & 53.34 & 49.06
        & 3.30 & 56.69 & 48.13 \\
        w/o Context
        & 3.55 & 54.16 & 49.65
        & \textbf{3.29} & 57.37 & 49.09 \\
        \textbf{Full MemoryVLN}
        & \textbf{3.52} & \textbf{54.76} & \textbf{50.26}
        & 3.33 & \textbf{57.84} & \textbf{50.42} \\
        \bottomrule
    \end{tabular}
\end{table*}

\paragraph{Effect of Long Sequence Compression.}
Removing Long Sequence Compression causes OOM on both benchmarks because historical visual tokens grow continuously with trajectory length. Thus, compression is not only an efficiency optimization but also necessary for long-horizon inference.

\paragraph{Effect of Ego-Pose Encoding.}
Removing pose encoding degrades most metrics on both benchmarks. Restoring it improves R2R-CE OS, SR, and SPL by 3.16, 2.01, and 1.01 points while reducing NE to 3.52, and improves RxR-CE SR, SPL, and nDTW by 0.60, 1.68, and 1.51 points. This confirms that ego-pose information helps disambiguate visually similar observations from different locations.

\paragraph{Effect of the In-Episode Bank.}
The In-Episode Bank raises R2R-CE SR and SPL from 53.34 and 49.06 to 54.76 and 50.26, respectively. On RxR-CE, it improves SR, SPL, and nDTW by 1.15, 2.29, and 1.69 points. These gains show that the short- and long-term pools preserve useful in-trajectory evidence for later decisions.

\begin{figure*}[!t]
    \centering

    \includegraphics[
        width=0.78\textwidth]{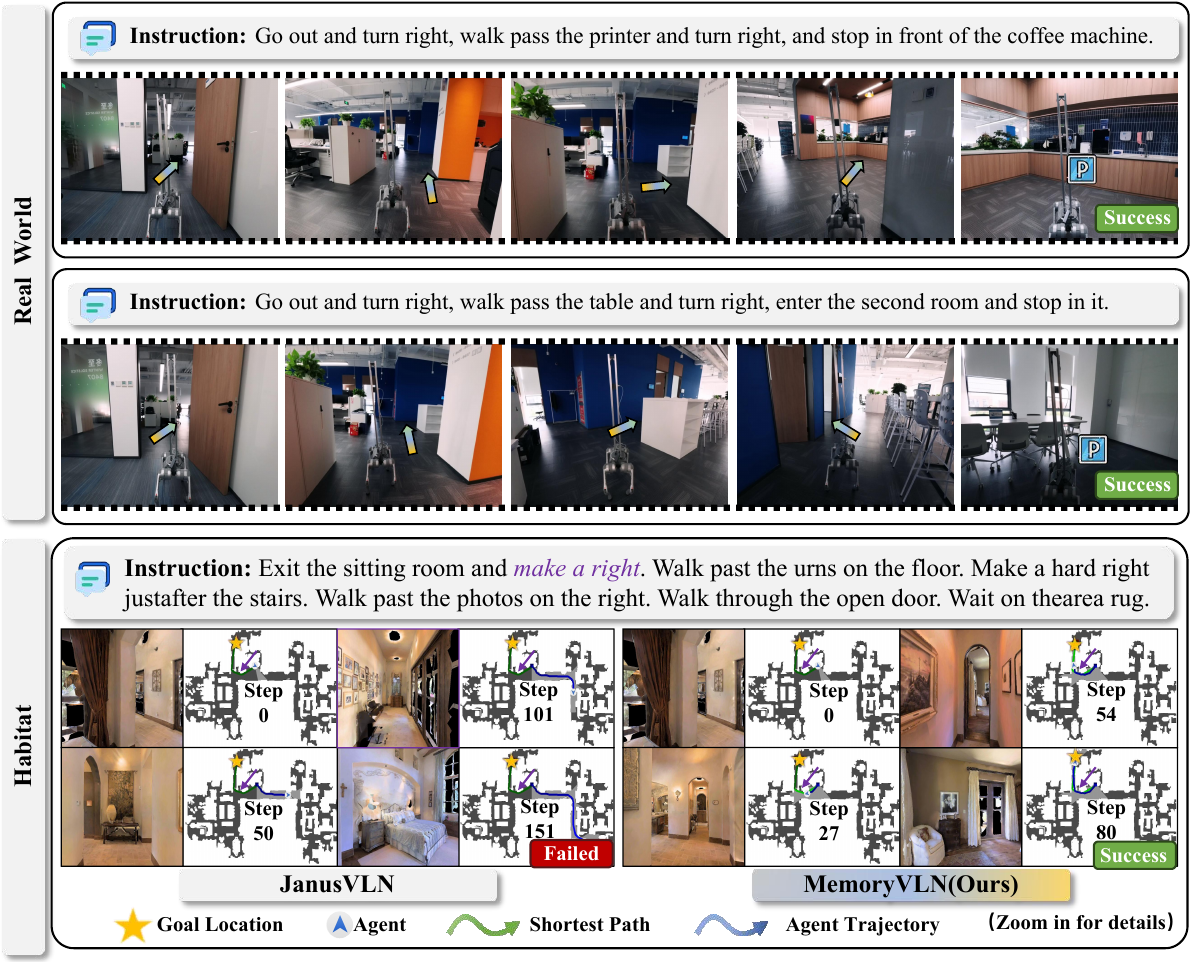}
    \captionof{figure}{Qualitative evaluation in Habitat simulation and a real indoor environment. Top: JanusVLN gradually deviates from the instructed route and fails, whereas MemoryVLN preserves historical and spatial context and reaches the target. Bottom: MemoryVLN successfully executes two real-world instructions with different routes and target locations.}
    \label{fig:exp_merge}

\end{figure*}
\paragraph{Effect of the Context-Level Bank.}
The Context-Level Bank improves R2R-CE OS, SR, and SPL by 1.52, 0.60, and 0.61 points and reduces NE from 3.55 to 3.52. On RxR-CE, it improves SR, SPL, and nDTW by 0.47, 1.33, and 0.56 points. Although RxR-CE NE increases slightly, the remaining gains indicate that cross-episode experience improves navigation success and trajectory quality.

\FloatBarrier

\subsection{Qualitative Analysis}

\autoref{fig:exp_merge} presents a qualitative evaluation of MemoryVLN in Habitat simulation and a real environment. In the long-horizon Habitat example shown at the top, JanusVLN loses instruction-relevant historical and spatial context, deviates from the intended route, and ultimately fails. In contrast, MemoryVLN maintains consistent historical-spatial information and reaches the target. The bottom two examples show MemoryVLN following different natural-language instructions toward targets in a real indoor environment. MemoryVLN completes both tasks, suggesting that the memory construction and management mechanisms learned in simulation can transfer to physical environments. Additional Habitat trajectories, real-world experiments, and qualitative results are provided in the supplementary material.

\FloatBarrier


\FloatBarrier
\section{Conclusion}
The main message of this paper is that the core bottleneck of long-horizon VLN is not simply context size, but the lack of Hierarchical Navigation Memory. MemoryVLN addresses this challenge through Long Sequence Compression, Ego-Aware Visuospatial Representation, and Memory Enhancement, transforming historical observations into navigation experience that is compact, spatially grounded, retrievable, and reusable. Results on R2R-CE and RxR-CE show that this hierarchical memory mechanism yields stable gains for long-horizon navigation. Future work may explore more persistent lifelong memory, more explicit map-like memory, and tighter integration between memory and planning.

\bibliographystyle{unsrtnat}
\bibliography{main}

\appendix

\begin{center}
  {\LARGE\bfseries Appendix}
\end{center}
\addcontentsline{toc}{section}{Appendix}
\noindent\textbf{Scope.} This supplementary document provides implementation and evaluation details. It is intended to be submitted as a standalone PDF together with the code-and-data package. It contains no links to external code repositories. Material that is essential to the main claims remains in the main paper.

\section{Implementation Details}

\subsection{Task, Data, and Evaluation Protocol}

We evaluate on the continuous-navigation versions of R2R and RxR, which use Matterport3D environments in Habitat. The reported main-table and ablation results use the val-unseen split. The model is trained with standard VLN trajectories only; no DAgger rollouts or additional navigation data are used for the main results. In accordance with the licenses of the underlying benchmarks, the submitted code package contains preprocessing scripts and path templates, rather than redistributed Matterport3D scenes or benchmark episode files.

At time step $t$, the agent receives an instruction and an egocentric RGB observation, and predicts one of \textsc{MoveForward}, \textsc{TurnLeft}, \textsc{TurnRight}, and \textsc{Stop}. Closed-loop evaluation is capped at 400 actions per episode. The In-Episode Bank is reset at each episode boundary, whereas the Context-Level Bank is retained only within the associated environment scope. This distinction prevents information from one scene from entering another scene's evaluation.

\subsection{Metrics}

Let $P_i$ be the predicted path for episode $i$, $R_i$ its reference path, $p_i$ and $r_i$ their lengths, $d(\cdot,\cdot)$ geodesic distance in the simulator, and $d_{\mathrm{th}}$ the benchmark success threshold. Navigation Error (NE) is the final geodesic distance to the goal,
\begin{equation}
\mathrm{NE}=\frac{1}{N}\sum_{i=1}^{N} d(P_i^{\mathrm{final}}, g_i).
\end{equation}
Success Rate (SR) measures whether the final state is within $d_{\mathrm{th}}$ of the goal; Oracle Success (OS) uses the closest state visited on the path:
\begin{equation}
\mathrm{SR}=\frac{1}{N}\sum_i \ind[d(P_i^{\mathrm{final}},g_i)\leq d_{\mathrm{th}}].
\end{equation}
\begin{equation}
\mathrm{OS}=\frac{1}{N}\sum_i \ind[\min_{x\in P_i} d(x,g_i)\leq d_{\mathrm{th}}].
\end{equation}
Success weighted by Path Length (SPL) jointly measures completion and path efficiency,
\begin{equation}
\mathrm{SPL}=\frac{1}{N}\sum_i \mathrm{SR}_i\frac{r_i}{\max(r_i,p_i)},
\end{equation}
and normalized Dynamic Time Warping (nDTW) measures trajectory similarity,
\begin{equation}
\mathrm{nDTW}=\frac{1}{N}\sum_i \exp\left(-\frac{\mathrm{DTW}(P_i,R_i)}{|R_i|d_{\mathrm{th}}}\right).
\end{equation}
We report these standard VLN-CE metrics because they separately capture final accuracy, whether the goal was ever reached, path efficiency, and adherence to the reference route.

\subsection{Implementation Configuration}

\method{} uses Qwen2.5-VL as the action-prediction backbone and keeps the VGGT geometry branch frozen. \autoref{tab:release-config} records the configuration used for the reported results. The submitted code package includes compact configuration files and a run manifest; model weights are omitted to satisfy the 50MB upload limit.

\begin{table*}[t]
\centering
\small
\setlength{\tabcolsep}{5pt}
\renewcommand{\arraystretch}{1.08}
\caption{Implementation values used for the reported results. The submitted \texttt{config.json} records the model and memory settings; an evaluation command may explicitly override these values.}
\label{tab:release-config}
\begin{tabular}{p{0.32\textwidth}p{0.16\textwidth}p{0.43\textwidth}}
\toprule
\textbf{Component} & \textbf{Value} & \textbf{Role}\\
\midrule
Action backbone & Qwen2.5-VL & Language-conditioned action prediction.\\
Geometry backbone & frozen VGGT & Produces geometric tokens; no VGGT parameters are updated.\\
Training / evaluation seed & 42 & Training and evaluation use this seed; evaluation seeds Python, NumPy, PyTorch CPU, and all CUDA devices.\\
Reported training runs & 1 & All reported results come from one training run with seed 42; no across-run variation is reported.\\
Training global batch & 64 & $64$ GPUs $\times$ per-device batch size $1$ $\times$ gradient accumulation $1$.\\
Compute platform & 64 $\times$ A100 80GB & 8 \texttt{ml.hpcpni2.28xlarge} instances; each has 112 vCPUs, 1960 GiB host memory, and \texttt{2.5.1-cu12.4.1-py3.9-ubuntu22.04}.\\
Maximum rollout length & 400 & Stops an episode after at most 400 navigation actions.\\
Frame-sampling chunk size & 8 & Chunk size for two-level Ebbinghaus-style sampling.\\
Inner / outer sampling scale & 5.0 / 3.0 & Controls within-chunk and cross-chunk retention.\\
Adaptive sampling start & 16 frames & Compression starts once the observed history exceeds this value.\\
Adaptive budget & $\max(\min(N,4+2\sqrt{N}),16)$ & Uses offset 4, square-root coefficient 2, and a minimum of 16 retained frames.\\
Recent tail & 1 frame & Keeps the latest observation in the sampling procedure.\\
Position encoding & enabled & Injects the 6D agent pose into the visual streams.\\
Frame-memory capacities & 4 short + 12 long & Total capacity of 16 frame-memory slots.\\
Scene-bank capacity & 300 & Maximum in-memory cross-episode entries.\\
Scene retrieval threshold & 0.6 & Cosine-similarity threshold for a retrieval hit.\\
Warm-start coefficient & 0.5 & Blend ratio for a retrieved scene prior.\\
Write-back confidence threshold & 0.4 & Prevents unreliable in-episode summaries from entering the scene bank.\\
\bottomrule
\end{tabular}
\end{table*}

\subsection{Software Environment and Randomness}

The released environment snapshot uses Python 3.9.23, PyTorch 2.6.0 with CUDA 12.4, Habitat-Sim 0.2.4, Habitat-Lab 0.2.4, Habitat-Baselines 0.2.4, Transformers 4.50.0, and FlashAttention 2.7.0.post2. The training entry point calls the Transformers \texttt{set\_seed} utility and the released launcher explicitly passes seed 42; the evaluation entry point likewise uses seed 42 for Python, NumPy, PyTorch, and all CUDA devices. Data-preparation tools use deterministic episode-derived seeds. The reported experiments run on eight \texttt{ml.hpcpni2.28xlarge} instances, for a total of 64 Tesla A100 80GB GPUs; each instance provides 112 vCPUs and 1960 GiB of host memory, using the \texttt{2.5.1-cu12.4.1-py3.9-ubuntu22.04} runtime image.

The reported training uses bfloat16 precision, AdamW, a cosine learning-rate schedule with a 3\% warmup ratio, weight decay of 0.01, a per-device batch size of 1, and gradient accumulation of 1 step. The resulting global batch size is 64 ($64$ GPUs $\times$ per-device batch size $1$ $\times$ gradient accumulation $1$). Full deterministic mode is intentionally not enabled, so the fixed seed supports comparable reruns rather than bitwise-identical training.

\section{More Ablation Studies}

The main paper reports aggregate values; the submitted package should additionally provide the episode-level evidence needed to recompute them. The code-and-data package additionally provides per-episode prediction JSON files and the aggregation entry point so that NE, OS, SR, SPL, and nDTW can be recomputed. This permits paired comparisons between \method{} and a baseline evaluated on the same episodes.

We pair the JanusVLN and MemoryVLN outputs by scene name and episode ID. The R2R files contain 1,839 matched episodes, while the RxR files contain 3,669 matched episodes. For the RxR SPL analysis, 99 pairs with non-finite values are excluded, leaving 3,570 valid episodes.

For binary SR, we use the exact McNemar test to compare discordant success/failure outcomes on the same episodes. For continuous SPL, we use a two-sided Wilcoxon signed-rank test. The confidence intervals are paired bootstrap 95\% intervals for the mean difference based on 20,000 resamples.

On R2R, MemoryVLN improves SR and SPL by 2.0 and 1.2 percentage points, although neither difference is statistically significant. On RxR, MemoryVLN improves SR and SPL by 6.2 and 5.2 percentage points. The RxR SR difference is significant under the exact McNemar test, and the SPL difference is significant under the two-sided Wilcoxon signed-rank test.

The line plots select three comparison stages: 26.3\% training data, 46.4\% training data, and the reported final results. At 26.3\% training data, JanusVLN obtains 38.2 SR and 30.8 SPL, whereas MemoryVLN obtains 38.6 SR and 33.1 SPL, improvements of 0.4 and 2.3 percentage points. At 46.4\% training data, JanusVLN obtains 44.4 SR and 37.6 SPL, whereas MemoryVLN obtains 48.6 SR and 44.7 SPL, improvements of 4.2 and 7.1 percentage points. For the reported final results, JanusVLN Base obtains 52.8 SR and 49.0 SPL, while MemoryVLN at step 4,745 obtains 54.8 SR and 50.3 SPL, improvements of 2.0 and 1.2 percentage points. 

\begin{figure*}[t]
\centering
\includegraphics[width=0.98\textwidth]{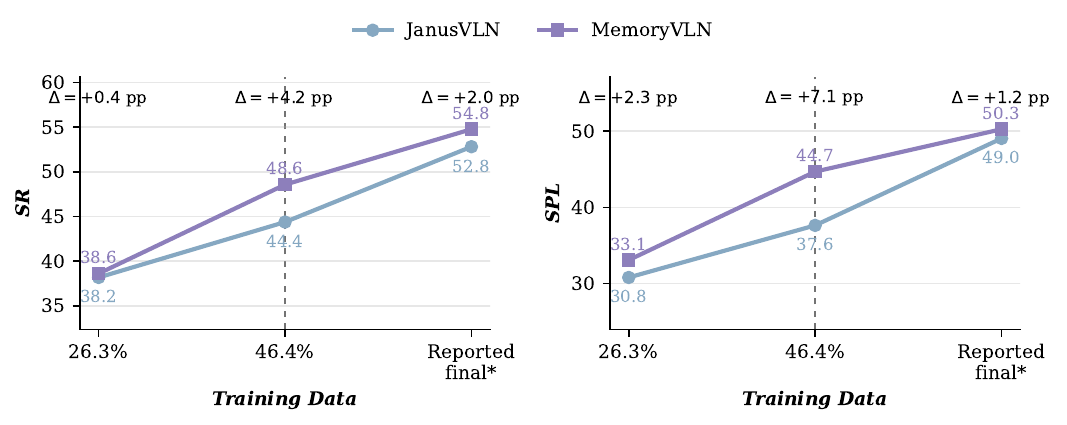}
\caption{Selected R2R val-unseen comparisons for SR and SPL as functions of training data. The dashed line marks 46.4\% training data, the last data-matched comparison point; the reported-final point to its right is marked with an asterisk because it is not training-data matched. Annotations report MemoryVLN minus JanusVLN in percentage points.}
\label{fig:selected-checkpoint-lines}
\end{figure*}

\begin{table*}[t]
\centering
\small
\setlength{\tabcolsep}{4pt}
\renewcommand{\arraystretch}{1.08}
\caption{Paired significance tests on the same evaluation episodes. Differences are MemoryVLN minus JanusVLN in percentage points; bootstrap intervals quantify episode-level uncertainty and do not represent across-run variability. For RxR SPL, 99 pairs with non-finite values are excluded, leaving 3,570 valid episode pairs.}
\label{tab:paired-significance}
\begin{tabular}{llrrrrllr}
\toprule
Dataset & Metric & $N$ & JanusVLN & MemoryVLN & $\Delta$ (pp) & 95\% CI & Test & $p$\\
\midrule
R2R & SR  & 1,839 & 52.8 & 54.8 & +2.0 & [-0.7, 4.6] & exact McNemar & 0.158\\
R2R & SPL & 1,839 & 49.0 & 50.3 & +1.2 & [-1.3, 3.7] & Wilcoxon signed-rank & 0.327\\
RxR & SR  & 3,669 & 51.0 & 57.2 & +6.2 & [4.5, 8.0] & exact McNemar & $2.71\times10^{-12}$\\
RxR & SPL & 3,570 & 43.5 & 48.7 & +5.2 & [3.8, 6.7] & Wilcoxon signed-rank & $1.27\times10^{-12}$\\
\bottomrule
\end{tabular}
\end{table*}

\section{Qualitative Analysis and Failure Cases}

\subsection{Interpretation of Memory Retrieval}

In long trajectories, the short-term pool preserves nearby observations while the long-term pool stores consolidated history. The Context-Level Bank is only reused when a query is sufficiently similar to an entry from the same environment. This separation is especially useful when a current observation is visually ambiguous: ego pose and retrieved history can distinguish similar-looking rooms that occupy different positions in the scene.

\subsection{Failure Case Analysis}
\label{sec:memoryvln-failure-analysis}

To examine the remaining limitations of MemoryVLN, we select four reproducible failed rollouts for qualitative analysis. Following the top-to-bottom order in the visualization, the cases cover room-level semantic branching, fine-grained final-doorway grounding, unsuccessful context reuse, and long-instruction progress loss. Navigation Error (NE) measures the final distance to the goal, while Oracle NE is the minimum goal distance reached anywhere along the predicted trajectory.

\begin{figure*}[t]
    \centering
    \includegraphics[height=0.85\textheight,keepaspectratio]{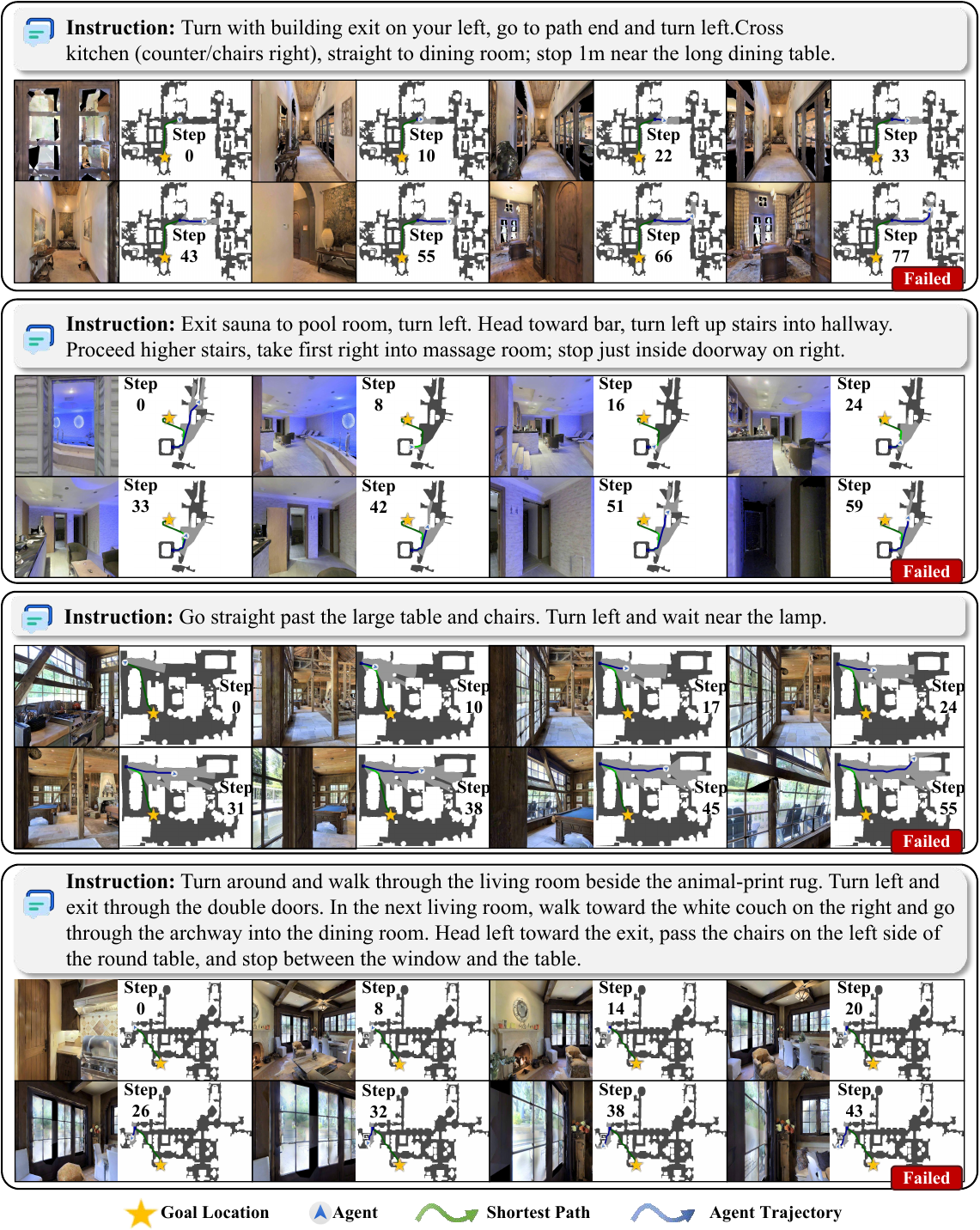}
    \caption{Representative MemoryVLN failure trajectories. From top to bottom, the rows illustrate room-level branch confusion, ordinal-sensitive final-doorway grounding, a landmark-turn error not corrected by context memory, and long-instruction progress loss. Each row pairs egocentric observations with a top-down map showing the goal, shortest path, and executed agent trajectory.}
    \label{fig:memoryvln-badcases}
\end{figure*}

\autoref{fig:memoryvln-badcases} pairs the egocentric observations of each case with its top-down trajectory, allowing the location and development of each navigation error to be inspected directly.

\begin{table*}[t]
\centering
\small
\setlength{\tabcolsep}{5pt}
\renewcommand{\arraystretch}{1.08}
\caption{Four representative failure cases in the same top-to-bottom order as \autoref{fig:memoryvln-badcases}. All four trajectories have zero Success, SPL, and Oracle Success.}
\label{tab:memoryvln-selected-failures}
\begin{tabular}{llrrrrl}
\toprule
Dataset & Scene / Episode & Steps & NE & Oracle NE & Coll. & Failure type \\
\midrule
R2R & \texttt{QUCTc6BB5sX / 577} & 78 & 28.61 & 14.13 & 0 & Semantic branch confusion \\
R2R & \texttt{x8F5xyUWy9e / 231} & 60 & 10.76 & 6.08 & 3 & Final-doorway grounding \\
R2R & \texttt{EU6Fwq7SyZv / 1206} & 56 & 10.28 & 5.82 & 0 & Failed-context reuse \\
RxR & \texttt{2azQ1b91cZZ / 5691} & 44 & 16.40 & 12.49 & 3 & Progress loss / early stop \\
\bottomrule
\end{tabular}
\end{table*}

\paragraph{Case 1: room-level branch and target-room confusion (R2R, \texttt{QUCTc6BB5sX / 577}).}
The instruction requires the agent to traverse the entrance and kitchen, continue into the dining room, and stop approximately one meter from the nearest end of the long dining table. MemoryVLN follows the entrance and kitchen portion correctly, but then selects the wrong room-level branch and terminates in a study- or office-like area rather than the dining room.

The final NE is 28.61 m and the Oracle NE remains 14.13 m, showing that the agent never approaches the target region during the rollout. The trajectory contains no collisions, so the failure is not primarily caused by low-level control or obstacle handling. Instead, the model fails to bind the later observations to the target semantics ``dining room'' and ``long dining table,'' and it does not verify whether the selected topological branch is consistent with the remaining instruction.

This case suggests adding explicit room-type verification at major branches and preserving unresolved target-landmark constraints until the requested room has been positively identified.

\paragraph{Case 2: fine-grained final-doorway grounding after ordered stair transitions (R2R, \texttt{x8F5xyUWy9e / 231}).}
The instruction requires the agent to leave the sauna, pass the pool and bar, climb two ordered stair segments, enter the first massage room on the right, and stop just inside the doorway. MemoryVLN completes the sauna, pool, bar, and part of the stair sequence, but fails to select the correct first-right entrance after the second stair segment and stops in an incorrect dark corridor or room.

MemoryVLN terminates after 60 steps, while the successful reference trajectory requires 82 steps, indicating that the final action sequence is abandoned before completion. Its final NE is 10.76 m and its Oracle NE is 6.08 m, showing that the agent reaches a relatively closer region but fails to complete the final doorway-level localization. Both trajectories contain approximately three collisions, while the reference still succeeds, so collision handling is not the main factor separating success from failure.

This error combines ordered-language grounding with fine-grained final localization: ``the first stairs,'' ``the next part of the stairs,'' ``the first right,'' and ``just inside the doorway on the right'' must be resolved jointly. Selecting a doorway from current appearance alone, without preserving the stair-transition count and ordinal constraints, is insufficient for this task. The model would benefit from explicit counters for stair and turn transitions, together with a final verification stage that checks doorway orientation, room semantics, and ordinal constraints before stopping.

\paragraph{Case 3: unsuccessful context write-back and reuse (R2R, \texttt{EU6Fwq7SyZv / 1206}).}
The target instruction asks the agent to move straight past the large table and chairs, turn left, and wait near the lamp before the stairs. MemoryVLN misses the required turn toward the lamp and stair region and instead continues along the upper corridor into the living and billiard-table area.

Before target episode 1206, episode 1204 fails with a final NE of 5.39 m, yet its context is still written into the Scene Bank. Episode 1206 subsequently retrieves this failed entry with a cosine similarity of 1.0000 and exhibits a similar early wrong-direction tendency. This trace exposes the absence of a reliability filter based on episode success, navigation error, or trajectory quality before context write-back and reuse.

The causal interpretation must remain conservative because disabling the Scene Bank reproduces the same 56-step failure and 10.28 m NE for episode 1206. Therefore, the Scene Bank is not the sole cause of the navigation error; the supported conclusion is that the base policy already makes a grounding mistake, while the current memory mechanism fails to filter that mistake and permits unsuccessful experience to remain reusable. A practical remedy is to gate write-back using success, final NE, path efficiency, and collision statistics, and to include an explicit reliability score when ranking retrieved entries.

\paragraph{Case 4: long-instruction progress loss and premature stopping (RxR, \texttt{2azQ1b91cZZ / 5691}).}
The RxR instruction contains a long sequence of subgoals involving a kitchen, a first living room, double doors, a second living room, an arched entrance, a round dining table, and a final window-side destination. A successful reference rollout requires 204 steps, whereas MemoryVLN stops after only 44 steps in an earlier living-room and doorway region.

MemoryVLN ends with an NE of 16.40 m and an Oracle NE of 12.49 m, indicating that it neither passes near the goal nor executes the latter portion of the instruction. Because the intermediate rooms also contain windows, doors, and furniture, local visual similarity may be mistaken for the final state ``between the window and the table.'' The central failure is that completed and pending subgoals are not represented explicitly enough, allowing the remaining instruction to be forgotten and the STOP action to be triggered prematurely.

This case motivates an explicit subgoal-progress state and a pre-STOP consistency check that verifies whether all later landmarks have been reached in the required order.

\paragraph{Overall diagnosis.}
Together, the four cases expose failures at four different levels: semantic-topological selection, fine-grained terminal grounding, memory quality control, and long-horizon progress tracking. They are not explained by a single collision or action-execution issue; instead, failure occurs when consistency breaks among instruction state, visual landmarks, topological location, and the STOP decision. Future improvements should therefore combine goal-conditioned branch verification, target-consistency validation before stopping, reliability-gated memory write-back, and explicit subgoal progress.

\subsection{More Qualitative Results}

\begin{figure*}[t]
    \centering
    \includegraphics[height=0.97\textheight,keepaspectratio]{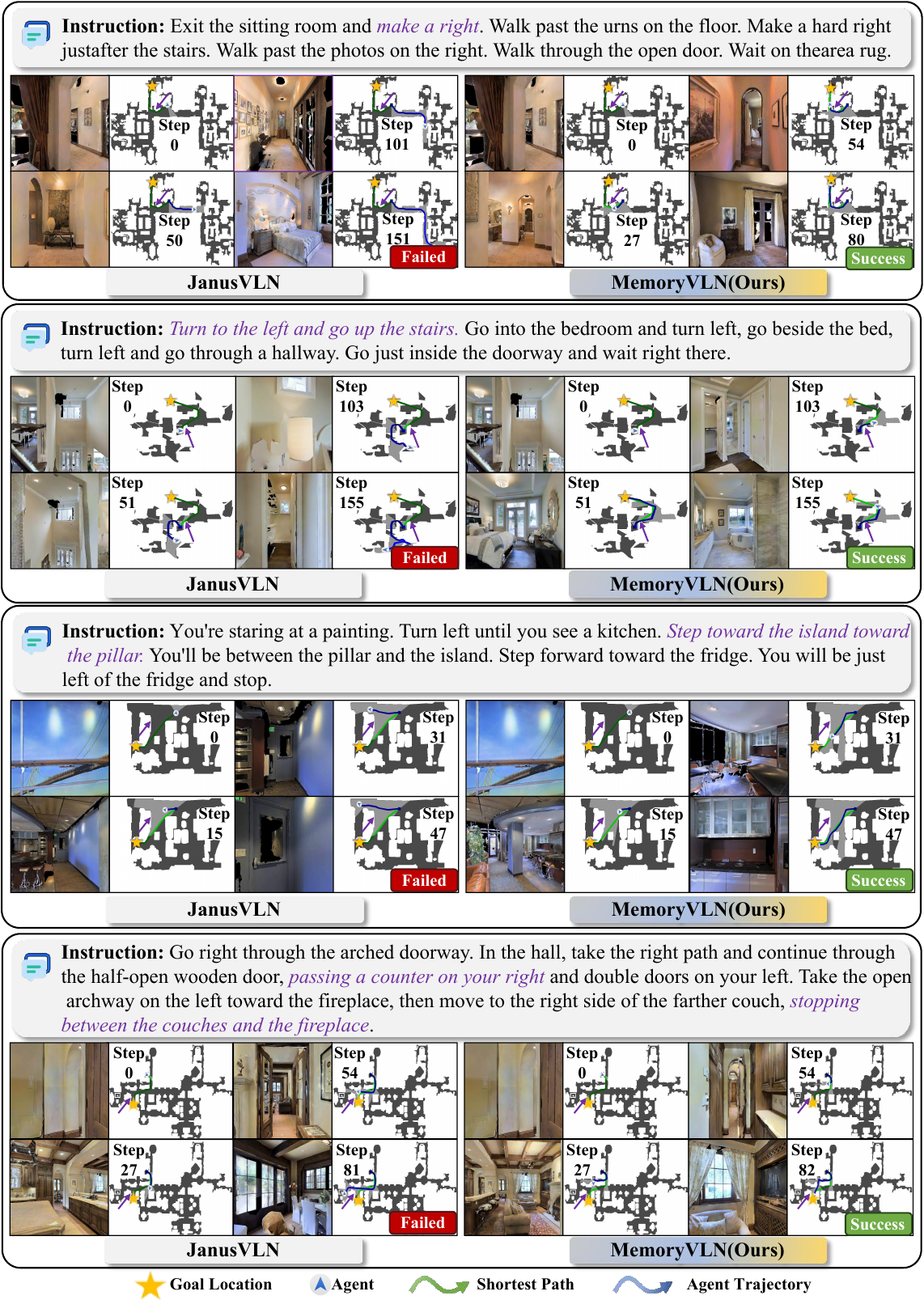}
    \caption{Qualitative comparison between JanusVLN and MemoryVLN on four Habitat episodes under identical instructions. Each example combines representative egocentric observations with top-down trajectories, where the yellow star marks the goal, the green curve marks the shortest path, and the blue curve marks the executed agent trajectory. JanusVLN fails in these examples, whereas MemoryVLN reaches the target region and stops successfully.}
    \label{fig:qualitative-habitat}
\end{figure*}

\paragraph{Habitat comparison.}
\autoref{fig:qualitative-habitat} compares JanusVLN and MemoryVLN side by side under four identical navigation instructions. Each example combines representative egocentric key frames with top-down trajectories, allowing the route choice, goal location, and stopping behavior to be inspected together. In these cases, JanusVLN exhibits route deviation, failure to reach the goal within the rollout, or an incorrect final stop, whereas MemoryVLN enters the target region and stops successfully.

The examples cover repeated turns, room-to-room motion, stair and corridor transitions, and long instructions requiring precise target localization. The trajectory overlays indicate that the improvement is not merely an earlier stop: MemoryVLN maintains a route that remains better directed toward the goal over the long horizon.

\begin{figure*}[t]
    \centering
    \includegraphics[height=0.9\textheight,keepaspectratio]{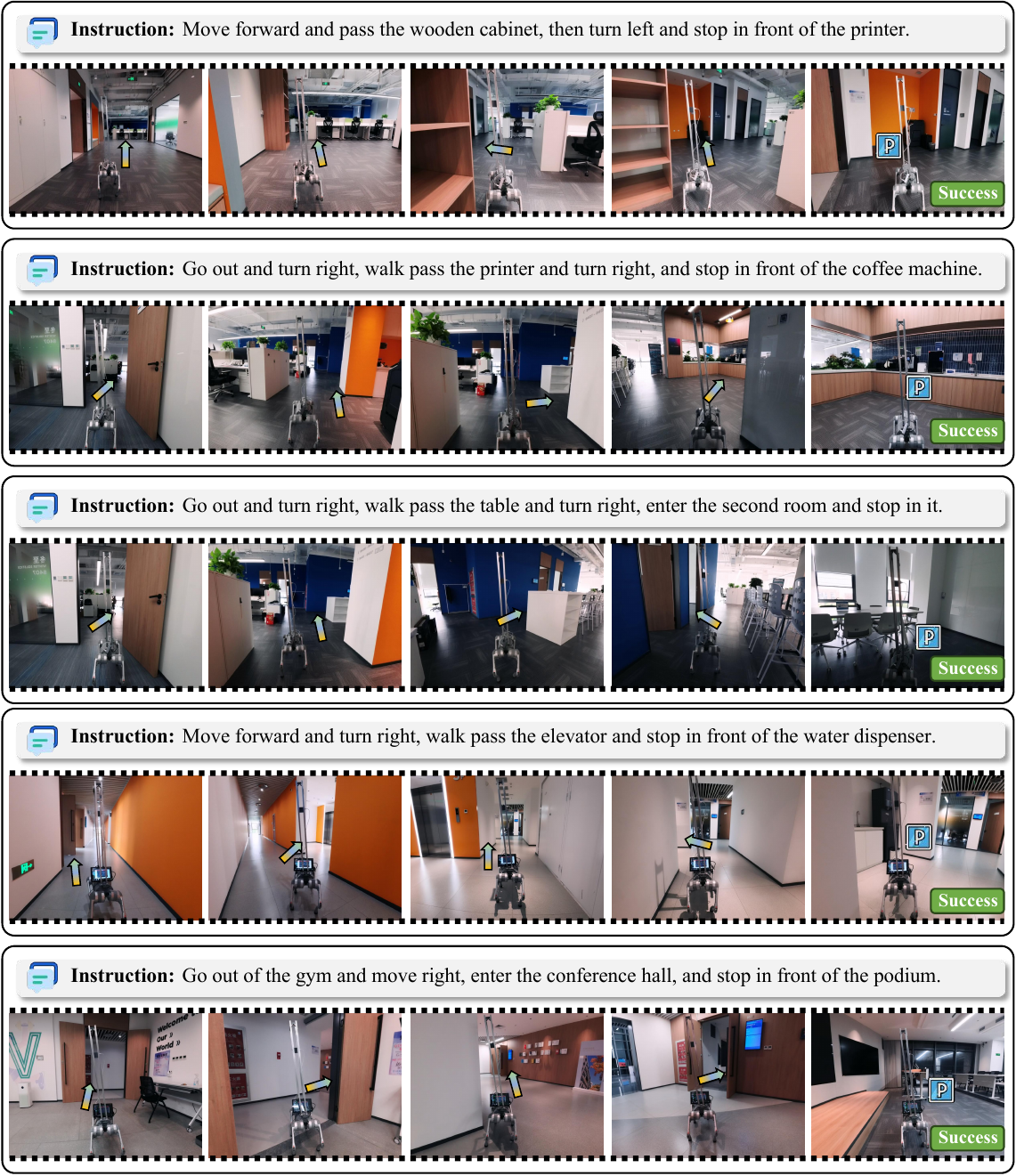}
    \caption{Qualitative real-world navigation results for five natural-language instructions using a Unitree GO2 quadruped in an indoor office environment. Navigation-policy inference is performed on a single NVIDIA GeForce RTX 4090 GPU. The egocentric sequences show the robot following corridor and doorway transitions, passing the specified landmarks, and stopping at the requested destination. The targets include a printer, a coffee machine, a designated room, a water dispenser, and a conference-room podium.}
    \label{fig:qualitative-real-world}
\end{figure*}

\paragraph{Real-world navigation.}
The real-world experiments use a Unitree GO2 quadruped as the mobile robotic platform. All navigation-policy inference is performed using a single NVIDIA GeForce RTX 4090 GPU. \autoref{fig:qualitative-real-world} shows MemoryVLN executing five natural-language instructions in a real office environment. The requested destinations are a printer, a coffee machine, a designated room, a water dispenser, and a conference-room podium. The egocentric sequences show the robot traversing doorways, corridors, and office landmarks before stopping at the requested target.

These real-world examples provide qualitative evidence that the navigation policy remains applicable beyond the visual appearance of Habitat simulation. The experiment is presented as a qualitative demonstration of real-robot deployment rather than a large-scale physical-world benchmark.

\end{document}